# Correcting Multi-focus Images via Simple Standard Deviation for Image Fusion

Firas A. Jassim
Management Information Systems Department, Irbid National University, Irbid 2600, Jordan
firasajil@yahoo.com

***Abstract***— Image fusion is one of the recent trends in image registration which is an essential field of image processing. The basic principle of this paper is to fuse multi-focus images using simple statistical standard deviation. Firstly, the simple standard deviation for the k×k window inside each of the multi-focus images was computed. The contribution in this paper came from the idea that the focused part inside an image had high details rather than the unfocused part. Hence, the dispersion between pixels inside the focused part is higher than the dispersion inside the unfocused part. Secondly, a simple comparison between the standard deviation for each k×k window in the multi-focus images could be computed. The highest standard deviation between all the computed standard deviations for the multi-focus images could be treated as the optimal that is to be placed in the fused image. The experimental visual results show that the proposed method produces very satisfactory results in spite of its simplicity.

***Index Terms***— Image fusion, Multi-focus, Multi-sensor, Standard deviation.

## I. INTRODUCTION

Image fusion is the process of integrating two or more images to construct one fused image which is highly informative. In accordance with the development of technology and inventing various image-capturing devices, it is not always possible to get a detailed image that contains all the required information. During the past few years, image fusion gets greater importance for many applications especially in remote sensing, computer vision, medical imaging, military applications, and microscopic imaging [27]. The multi-sensor data in the field of remote sensing, medical imaging and machine vision may have multiple images of the same scene providing different information. In machine vision, due to the limited depth-of-focus of optical lenses in charge-coupled device (CCD), it is not possible to have a single image that contains all the information of objects in the image. Sometimes, a complete picture may not be always feasible since optical lenses of imaging sensor especially with long focal lengths, only have a limited depth of field [9]. The basic objective image fusion is to extract the required information of an image that was captured from various sources and sensors. After that, fuse these images into single fused (composite) enhanced image [14]. In fact, the images that are captured by usual camera sensors have uneven characteristics and outfit different information. Therefore, the essential usefulness of image fusion is its important ability to realize features that can not be realized with traditional type of sensors which leads to enhancing human visual perception. In fact, the fused image contains greater data content for the scene than any one of the individual image sources alone [7]. Image fusion helps to get an image with all the information. The fusion process must preserve all relevant information in the fused image.

This paper discusses the fusion of multi-focus images using the simple statistical standard deviation in k×k window size. Section II gives a review of the recent work and produces variety of different approaches reached. The proposed technique for fusing images had been presented in section III. The experimental results and performance assessments are discussed in section IV. At the end, the main conclusions of this paper have been discussed in section V.

## II. IMAGE FUSION PRELIMINARIES

The essential seed of image fusion goes back to the fifties and sixties of the last century. It was the first time to search for practical methods of compositing images from different sensors to construct a composite image which could be used to better coincide natural images. Image fusion filed consists of many terms such as merging, combination, integration, etc [26].

Nowadays, there are two approaches for image fusion, namely spatial fusion and transform fusion. Currently, pixel based methods are the most used technique of image fusion where a composite image has to be synthesized from several input images [1][21]. A new multi-focus image fusion algorithm, which is on the basis of the ratio of blurred and original image intensities, was proposed by [19]. A generic categorization is to consider a process at signal, pixel, or feature and symbolic levels [15]. An application of artificial neural networks to solve multi-focus image fusion problem was discussed by [14]. Moreover, the implementation of graph cuts into image fusion to support was proposed by [16]. According to [23], a compressive sensing image fusion had been discussed. A wavelet-based image fusion tutorial must not be forgotten and could be found in [18]. A complex wavelets and its application to image fusion can be found in [11]. Image fusion using principle components analysis (PCA) in Compressive sampling domain could be found [28]. A new Intensity-Hue-Saturation (HIS) technique to Image Fusion with a Tradeoff Parameter was discussed by [3]. An implementation of Ripplet transform into medical images was discussed by [6]. An excellent comparative analysis of image fusion methods can be found in [26]. A comparison of different image fusion techniques for individual tree crown identification using quickbird images [20].

Here, in this paper, in order to compare the proposed technique with some common techniques in image fusion, therefore; we have used the wavelet and principle components analysis (PCA) as the most two widely implemented methods in image fusion [28]. The wavelet transform has become a very useful tool for image fusion. It has been found that wavelet-based fusion techniques outperform the standard fusion techniques in spatial and spectral quality, especially in minimizing color distortion. Schemes that combine the standard methods (HIS or PCA) with wavelet transforms produce superior results than either standard methods or simple wavelet-based methods alone. However, the tradeoff is higher complexity and cost [18]. On the other hand, principal component analysis is a statistical analysis for dimension reduction. It basically projects data from its original space to its eigenspace to increase the variance and reduce the covariance by retaining the components corresponding to the largest eigenvalues and discarding other components. PCA helps to reduce redundant information and highlight the components with biggest influence [26]. Furthermore, many tinny and big modifications on the known wavelet method were researched and discussed by many authors. On of these contributions the one that was introduced for image fusion named wavelet packet by [2]. Another contribution on wavelet called Dual Tree Complex Wavelet Transform (DT-CWT) can be found in [10][12]. Also, a curvelet image fusion was presented by [4][17]. Furthermore, many other fusion methods like contourlet by [8][22], and Non-subsampled Contourlet Transform (NSCT) which can be found in [5].

### III. PORPOSED TECHNIQUE

The proposed technique in this paper was based on the implementation of the simple standard deviation for each of the fused and input images. At the beginning, a brief description about statistical standard deviation must be introduced. Actually, the standard deviation is a measure of how spreads out numbers are. Moreover, it is the measure of the dispersion of a set of data from its mean. The more spread apart the data, the higher the deviation. The standard deviation has proven to be an extremely useful measure of spread in part because it is mathematically tractable. In this paper, the essential contribution that was used as a novel image fusion technique was based on the fact that the standard deviation in the focused part inside an image had high details rather than its identical part in the unfocused image. Hence, the dispersion between pixels inside the focused part is higher than the dispersion inside the unfocused part. Hence, the standard deviation could be computed for each k×k window inside all the input multi-focus images. The higher value of the standard deviation between all the computed standard deviations in all the input images may be treated as the optimal standard deviation. According to Fig. 1, the standard deviation for an arbitrary 2×2 window size was computed to show that the higher standard deviation value in the four input images is the image that has more details. Statistically speaking, the computed standard deviations for the four input images were 0.5, 1.7078, 3.304, and 2.63, respectively; for figures (1.a), (1.b), (1.c), and (1.d). Therefore, the higher value of the standard deviation is (3.304) which belong to the third image (Fig. 1.c) which has sharp edges rather than the other images with blurred or semi-blurred edges. Consequently, the k×k window with the high standard deviation value is recommended to be the optimal window and that is to be placed in the fused image from all the input images. However, the suitable value of the window size k is preferred to be as small as possible. Therefore, in this article, the best recommended k that is to be gives best results is 2.

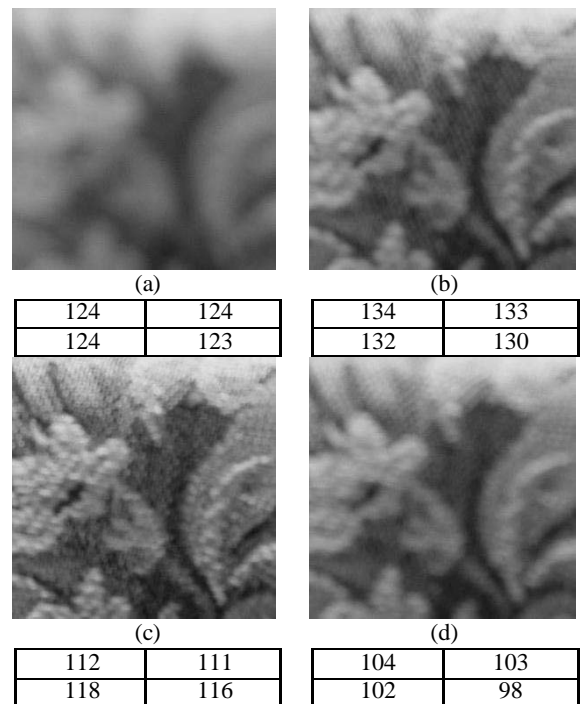

Figure 1. Four multi-focus images (a), (b), (c) and (d), each with its own standard deviation

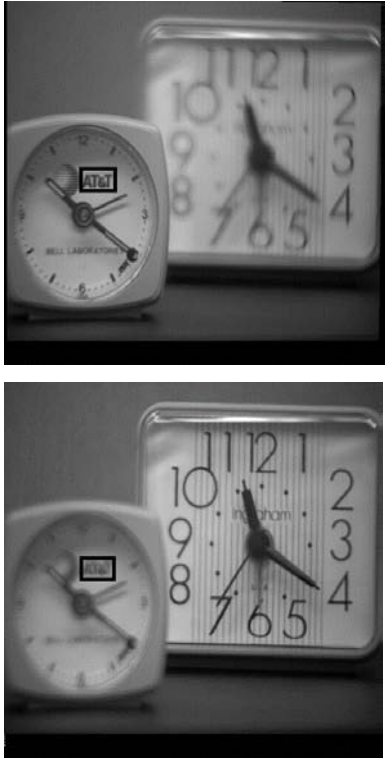

Figure 2. Blurred Block from Clock (Blur)

| 188 | 188 | 186 | 185 | 186 | 184 | 182 | 180 | 179 | 178 |
| 188 | 187 | 186 | 185 | 184 | 183 | 182 | 182 | 181 | 179 |
| 187 | 188 | 187 | 186 | 184 | 183 | 182 | 182 | 181 | 179 |
| 188 | 188 | 187 | 187 | 186 | 184 | 182 | 180 | 179 | 178 |
| 188 | 187 | 187 | 186 | 185 | 184 | 181 | 179 | 176 | 175 |
| 187 | 186 | 185 | 184 | 182 | 182 | 180 | 178 | 174 | 172 |
| 186 | 185 | 183 | 182 | 181 | 180 | 178 | 176 | 173 | 171 |
| 186 | 185 | 183 | 181 | 181 | 180 | 177 | 174 | 172 | 170 |
| 184 | 183 | 182 | 181 | 179 | 177 | 174 | 173 | 171 | 168 |
| 180 | 180 | 179 | 178 | 177 | 174 | 171 | 170 | 167 | 164 |

Figure 3. Original Block from Clock (Sharp)

| 189 | 189 | 188 | 186 | 185 | 183 | 181 | 182 | 183 | 182 |
| 190 | 190 | 190 | 189 | 186 | 183 | 181 | 182 | 182 | 181 |
| 188 | 188 | 188 | 188 | 187 | 184 | 182 | 181 | 181 | 181 |
| 188 | 187 | 187 | 187 | 188 | 186 | 183 | 182 | 182 | 182 |
| 189 | 188 | 189 | 189 | 187 | 186 | 184 | 182 | 182 | 182 |
| 188 | 187 | 187 | 188 | 186 | 185 | 184 | 182 | 181 | 181 |
| 187 | 186 | 185 | 185 | 185 | 185 | 184 | 183 | 182 | 181 |
| 188 | 187 | 187 | 187 | 185 | 186 | 186 | 184 | 182 | 181 |
| 186 | 187 | 188 | 188 | 187 | 183 | 182 | 183 | 182 | 181 |
| 184 | 184 | 185 | 186 | 185 | 181 | 182 | 182 | 179 | 176 |

Figure 4. Blurred Block from Clock (Blur)

As another example, for the matter of changing the window size, a 10×10 window size was used from clock multi-focus image, Fig. 2. The computed standard deviations for the 10×10 window size were (3.2489) and (1.8803), for the sharp and blurred images (for the small clock), respectively. The 10×10 window size was indicated by wide dark borders. Once again it is clear that the sharp edges image has higher standard deviation than the blurred image. The 10×10 windows sizes for the sharp and the blurred images were presented in figures (3) and (4), respectively.

## IV. EXPERIMENTAL RESULTS

In this section, experimental results that support the proposed technique in this paper were presented and discussed. Several standard test images were used as input images and the resulted fused images were obtained according to the proposed technique by implementing the application of the highest standard deviation value. Obviously, the ocular results show that the proposed technique produces quite satisfactory results compared with other methods like wavelet transform and principle components analysis (PCA). Practically, there are many performance measures used in image fusion techniques like Root Mean Square Error (RMSE), Peak Signal-to-Noise ratio (PSNR), Image Quality Index (IQI) [24], and structure similarity index (SSIM) [25][29]. Here, in this paper, the SSIM measure was implemented because of its high confidentiality [13].

TABLE 1 COMPARISON OF SSIM BETWEEN WAVELET, PCA, AND THE PROPOSED

|  | Wavelet | PCA | Proposed |
| --- | --- | --- | --- |
| Lena | 0.8278 | 0.8274 | 0.7260 |
| Clocks | 0.9224 | 0.9222 | 0.8807 |
| Peppers (V) | 0.8168 | 0.8166 | 0.7051 |
| Peppers(D) | 0.8193 | 0.8190 | 0.7079 |

In accordance with Table (1), the SSIM values for the proposed technique are less than those for both wavelet and PCA. It seems that the structure similarity index in both wavelet and PCA is higher than the proposed fusion technique. At this point, it must be mentioned that the SSIM shows the similarity between the fused and the input images according to the original structure. Since the blurring size used in this paper for all test images is approximately half of the input images, then SSIM shows the similarity according to both blurred and sharp parts of the input images. Therefore, the smaller value of SSIM for the proposed technique does not mean that it is not good. But it exactly means that the degree of similarity for the fused image constructed by the proposed technique is less than the degree of similarity for both the wavelet and PCA. This contribution can be consolidated by the ocular results in Figures (5) to (8). Actually, the visual results obtained by the proposed technique have more sharpening details than other methods (wavelet and PCA).

As another performance measure, the PSNR was used to test the goodness of the constructed image. Obviously, the PSNR measure can not be computed unless the original image must be present. This can not be done when using SSIM because the original image may not be always available because the multi-focused images are obtained in different zooming times and angles for the same scene. In fact, there is no perfect image that can give the ability to be compared with. Hence, in this paper, the PSNR can not be evaluated for all the test images. Therefore, three test images have their original images present which are Lena, Peppers (V), and Peppers (D), where V stands for Vertical and D for diagonal blurring direction.

TABLE 2 COMPARISON OF PSNR BETWEEN WAVELET, PCA, AND THE PROPOSED

|  | Wavelet | PCA | Proposed |
|---|---|---|---|
| Lena | 5.1918 | 5.1955 | 5.7523 |
| Peppers (V) | 5.0185 | 4.9843 | 5.2417 |
| Peppers(D) | 4.6687 | 4.6724 | 5.2405 |

Clearly, according to Table (2), the PSNR values for the proposed technique are higher (in small absolute difference) than the PSNR values for both the wavelet and PCA for the fused images. Actually, this means that the differences between the constructed image using the proposed technique and the original image are less than those of the wavelet and PCA. This result supports the contribution that was stated previously which is that the proposed technique produces quite satisfactory results according to its similar common fusion methods such as wavelet and PCA.

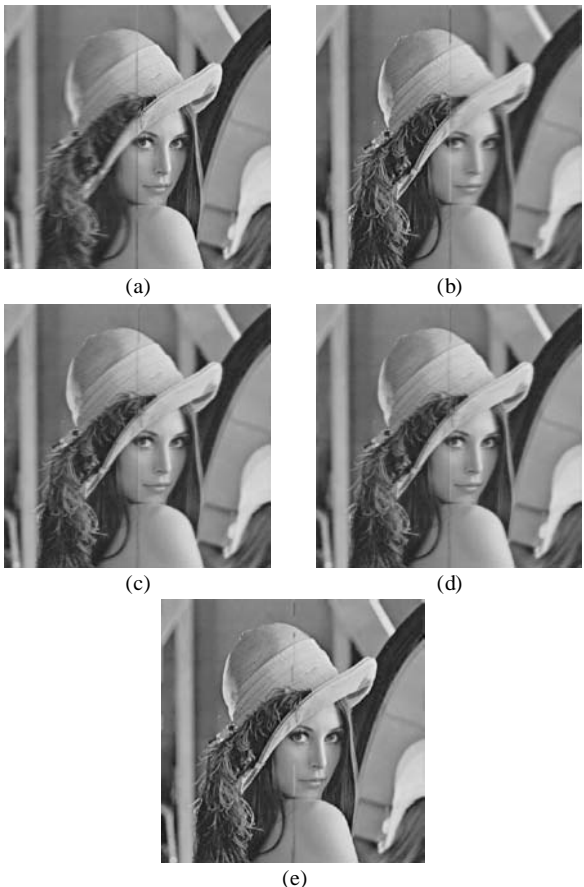

Figure 5 (a) Blurred Image (left) (b) Blurred image (right) (c) Wavelet (d) PCA (e) Proposed

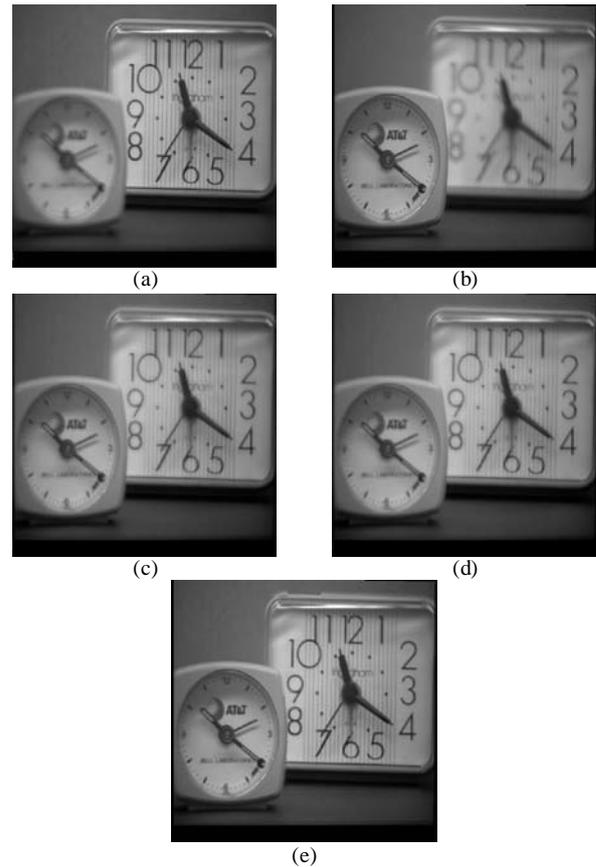

Figure 6 (a) Blurred Image (left) (b) Blurred image (right) (c) Wavelet (d) PCA (e) Proposed

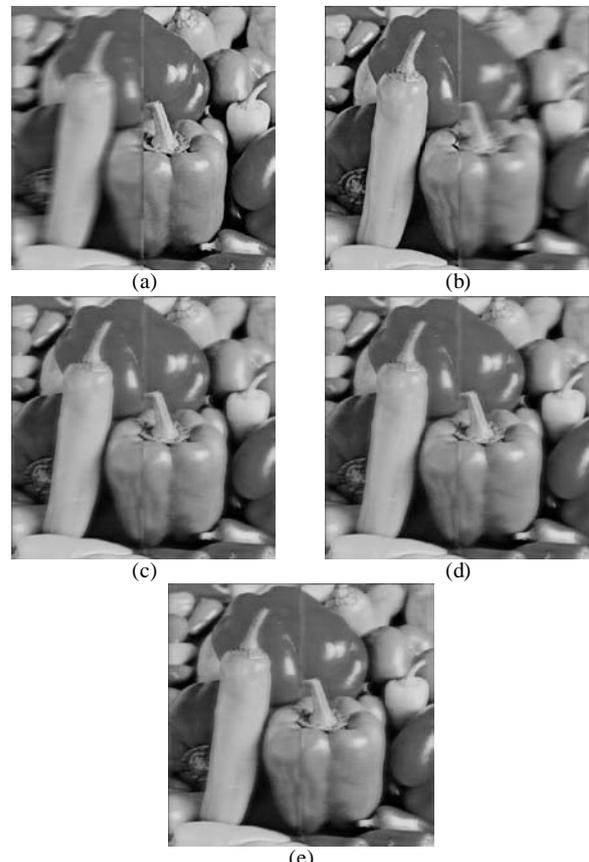

Figure 7 (a) Blurred Image (left) (b) Blurred image (right) (c) Wavelet (d) PCA (e) Proposed

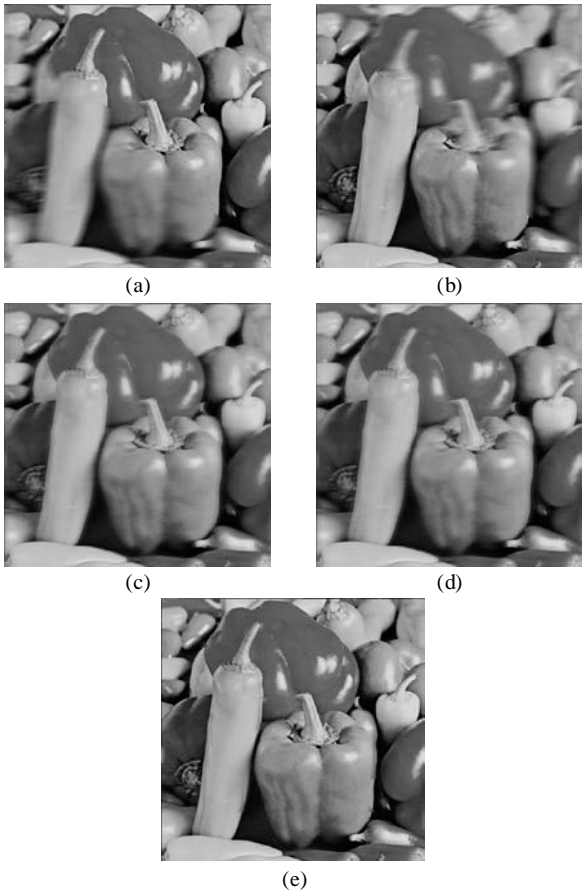

Figure 8 (a) Blurred Image (left) (b) Blurred image (right) (c) Wavelet (d) PCA (e) Proposed

## V. CONCLUSION

The main and fundamental conclusion that comes from the proposed technique is that its simplicity and performance compared to other complicated methods that produce almost the same results by using highly complicated mathematical formulas and time consuming transformations. The only disadvantage of the proposed technique is that the fused image is not always perfect. Hence, there must be a condition for the selection criterion for the best standard deviation between several standard deviations. This may be solved by future work. Further, the number of input images that was used in this article is two input images only. But this number can be generalized for more than two input images and that is the reason for the name multi-focused image fusion technique.

The results obtained by the highest value of the standard deviation technique has proven it's adequacy and efficiency in constructing a fused image from multi-focus images which were taken by different zooming process and times in the camera sensor. Finally, visual effect and statistical parameters indicate that the performance of our new method is better its competitors of the filed.

Additionally, one of the best recommendations that can be treated as a future work is the implementation of the proposed technique on color images than the gray scale images that were used in this article. Actually, this may facilitate the way for applying the proposed technique into video stream.

**Firas A. Jassim**, male, was born in Baghdad, Iraq, in 1974. He received the B.S. and M.S. degrees in Applied Mathematics and Computer Applications from Al-Nahrain University, Baghdad, Iraq, in 1997 and 1999, respectively, and the Ph.D. degree in Computer Information Systems (CIS) from the Arab University for Banking and Financial Sciences, Amman, Jordan, in 2012. In 2012, he joined the faculty of business administration, department of management information systems, Irbid National University, Irbid, Jordan, where he is currently an assistance professor. His current research interests are image compression, image interpolation, image segmentation, image enhancement, image fusion and simulation.